\documentclass[11pt,a4paper]{article}
\usepackage[hyperref]{acl2019}
\usepackage{times}
\usepackage{latexsym}
\usepackage{graphicx}
\usepackage{xspace}
\usepackage{url}
\usepackage{enumitem}

\usepackage{amsmath}
\usepackage{pgfplots}
\pgfplotsset{compat=1.14}

\newcommand{\bigruatt}{\textsc{bigru-att}\xspace}

\newcommand{\bigru}{\textsc{bigru}\xspace}
\newcommand{\gru}{\textsc{gru}\xspace}
\newcommand{\wordvec}{\textsc{word2vec}\xspace}

\newcommand{\rnn}{\textsc{rnn}\xspace}
\newcommand{\cnn}{\textsc{cnn}\xspace}
\newcommand{\han}{\textsc{han}\xspace}

\newcommand{\lmtc}{\textsc{lmtc}\xspace}

\newcommand{\zacnn}{\textsc{zacnn}\xspace}

\newcommand{\zagru}{\textsc{zagru}\xspace}
\newcommand{\lwan}{\textsc{lwan}\xspace}

\newcommand{\lwancnn}{\textsc{cnn-lwan}\xspace}
\newcommand{\zlwancnn}{\textsc{zero-cnn-lwan}\xspace}
\newcommand{\lwangru}{\textsc{bigru-lwan}\xspace}
\newcommand{\lwangrulv}{\textsc{bigru-lwan-l2v}}
\newcommand{\lwangruelmo}{\textsc{bigru-lwan-elmo}}
\newcommand{\zlwangru}{\textsc{zero-bigru-lwan}\xspace}

\newcommand{\glove}{\textsc{glove}\xspace}
\newcommand{\lawvec}{\textsc{l2v}\xspace}
\newcommand{\elmo}{\textsc{elmo}\xspace}
\newcommand{\bert}{\textsc{bert}\xspace}
\newcommand{\bertbase}{\textsc{bert-base}\xspace}

\newcommand{\newdata}{\textsc{eurlex57k}\xspace}

\newcommand{\eurovoc}{\textsc{eurovoc}\xspace}
\newcommand{\eu}{\textsc{eu}\xspace}
\newcommand{\rcv}{\textsc{rcv1}\xspace}

\newcommand{\eurlex}{\textsc{eur-lex}\xspace}

\newcommand{\mimicii}{\textsc{mimic-ii}\xspace}
\newcommand{\mimiciii}{\textsc{mimic-iii}\xspace}
\newcommand{\tfidf}{\textsc{tf-idf}\xspace}

\newcommand{\gpu}{\textsc{gpu}\xspace}

\aclfinalcopy 

\title{Large-Scale Multi-Label Text Classification on EU Legislation}

\author{Ilias Chalkidis \qquad Manos Fergadiotis \qquad Prodromos Malakasiotis \\ \textbf{Ion Androutsopoulos} \\ Department of Informatics, Athens University of Economics and Business, Greece \\ 
{\tt {\normalsize[ihalk,fergadiotis,rulller,ion]@aueb.gr}}}

\date{}

\begin{document}
\maketitle
\begin{abstract}
We consider Large-Scale Multi-Label Text Classification (\lmtc) in the legal domain. We release a new dataset of 57k legislative documents from \eurlex, annotated with $\mathtt{\sim}$4.3k \eurovoc labels, which is suitable for \lmtc, few- and zero-shot learning. Experimenting with several neural classifiers, we show that \bigru{s} with label-wise attention perform better than other current state of the art methods. Domain-specific \wordvec and context-sensitive \elmo embeddings further improve performance. We also find that considering only particular zones of the documents is sufficient. This allows us to bypass \bert's maximum text length limit and fine-tune \bert, obtaining the best results in all but zero-shot learning cases. 
\end{abstract}

\section{Introduction}
Large-scale multi-label text classification (\lmtc) is the task of assigning to each document all the relevant labels from a large set, typically containing thousands of labels (classes). Applications include building web directories \citep{Partalas2015LSHTCAB}, labeling scientific publications with concepts from ontologies \cite{Tsatsaronis2015}, assigning diagnostic and procedure labels to medical records \cite{Mullenbach2018,Rios2018-2}. We focus on legal text processing, an emerging \textsc{nlp} field with many applications (e.g., legal judgment \cite{Nallapati2008,Aletras2016}, contract element extraction \cite{Chalkidis2017}, obligation extraction \cite{Chalkidis2018b}), but limited publicly available resources.

Our first contribution is a new publicly available legal \lmtc dataset, dubbed \newdata, containing 57k English \eu legislative documents from the \eurlex portal, tagged with $\mathtt{\sim}$4.3k labels (concepts) from the European Vocabulary (\eurovoc).\footnote{See \url{https://eur-lex.europa.eu/} for \eurlex, and \url{https://publications.europa.eu/en/web/eu-vocabularies} for \eurovoc.} \eurovoc contains approx.\ 7k labels, but most of them are rarely used, hence they are under-represented (or  absent) in \newdata, making the dataset also appropriate for few- and zero-shot learning. \newdata can be viewed as an improved version of the dataset released by \citet{Mencia2007}, which has been widely used in \lmtc research, but is less than half the size of \newdata (19.6k documents, 4k \eurovoc labels) and more than ten years old. 

As a second contribution, we experiment with several neural classifiers on \newdata, including the Label-Wise Attention Network of \citet{Mullenbach2018}, called \lwancnn here, which was reported to achieve state of the art performance in \lmtc on medical records. We show that a simpler \textsc{bigru} with self-attention \cite{Xu2015} outperforms \lwancnn by a wide margin on \newdata. However, by replacing the \cnn encoder of \lwancnn with a \bigru, we obtain even better results on \newdata. Domain-specific \wordvec \cite{Mikolov2013} and context-sensitive \elmo embeddings \cite{Peters2018} yield further improvements. We thus establish strong baselines for \newdata.

As a third contribution, we investigate which zones of the documents are more informative on \newdata, showing that considering only the title and recitals of each document leads to almost the same performance as considering the full document. This allows us to bypass \bert's \cite{bert} maximum text length limit and fine-tune \bert, obtaining the best results for all but zero-shot learning labels. To our knowledge, this is the first application of \bert to an \lmtc task, which provides further evidence of the superiority of pretrained language models with task-specific fine-tuning, and establishes an even stronger baseline for \newdata and \lmtc in general.

\section{Related Work}
\label{sec:relatedwork}

You et al.\ \shortcite{You2018} explored \rnn-based methods with self-attention on five \lmtc datasets that had also been considered by \citet{Liu2017}, namely \rcv \cite{Lewis2004}, Amazon-13K, \cite{McAuley2013}, Wiki-30K and Wiki-500K \cite{Zubiaga2012}, as well as the previous \eurlex dataset \cite{Mencia2007}, reporting that attention-based \rnn{s} produced the best results overall (4 out of 5 datasets). 

\citet{Mullenbach2018} investigated the use of label-wise attention in \lmtc for medical code prediction on the \mimicii and \mimiciii datasets \cite{Johnson2017}. Their best method, Convolutional Attention for Multi-Label Classification, called \lwancnn here, employs one attention head per label and was shown to outperform weak baselines, namely logistic regression, plain \bigru{s}, \cnn{s} with a single convolution layer. 

\citet{Rios2018-2} consider few- and zero-shot learning on the \textsc{mimic} datasets. They propose Zero-shot Attentive \cnn, called \zlwancnn here, a method similar to \lwancnn, which also exploits label descriptors. Although \zlwancnn did not outperform \lwancnn overall on \mimicii and \mimiciii, it had much improved results in few-shot and zero-shot learning, among other variations of \zlwancnn that exploit the hierarchical relations of the labels with graph convolutions.

We note that the label-wise attention methods of \citet{Mullenbach2018} and \citet{Rios2018-2} were not compared to strong generic text classification baselines, such as attention-based \rnn{s} \cite{You2018} or Hierarchical Attention Networks (\han{s}) \cite{Yang2016}, which we investigate below.

\section{The New Dataset}
\label{sec:dataset}

As already noted, \newdata contains 57k legislative documents from \eurlex\footnote{Our dataset is available at \url{http://nlp.cs.aueb.gr/software_and_datasets/EURLEX57K}, with permission of reuse under European Union\copyright, \url{https://eur-lex.europa.eu}, 1998--2019.} with an average length of 727 words (Table~\ref{tab:dataset}).\footnote{See Appendix~\ref{app:dataset} for more statistics.} Each document contains four major zones: the \emph{header}, which includes the title and name of the legal body enforcing the legal act; the \emph{recitals}, which are legal background references; the \emph{main body}, usually organized in articles; and the \emph{attachments} (e.g., appendices, annexes).

\begin{table}[t]
\centering
\footnotesize
\begin{tabular}{lcccc}
  Subset & Documents ($D$) & Words/$D$ & Labels/$D$ \\
\hline
  Train & 45,000 & 729  & 5\\
  Dev. & 6,000 & 714  & 5 \\
  Test & 6,000 & 725  & 5\\\hline
  Total & 57,000 & 727 & 5\\
  \hline
\end{tabular}
\caption{Statistics of the \textsc{\eurlex} dataset.}
\label{tab:dataset}
\end{table}

Some of the \lmtc methods we consider need to be fed with documents split into smaller units. These are often sentences, but in our experiments they are \emph{sections}, thus we preprocessed the raw text,  respectively. We treat the header, the recitals zone, each article of the main body, and the attachments as separate sections. 

All the documents of the dataset have been annotated by the Publications Office of \eu\footnote{See \url{https://publications.europa.eu/en}.} with multiple concepts from \eurovoc. 
While \eurovoc includes approx.\ 7k concepts (labels), only 4,271 (59.31\%) are present in \newdata, from which only 2,049 (47.97\%) have been assigned to more than 10 documents. Similar distributions were reported by \citet{Rios2018-2} for the \textsc{mimic} datasets.
We split \newdata into training (45k documents), development (6k), and test subsets (6k). We also divide the 4,271 labels into \emph{frequent} (746 labels), \emph{few-shot} (3,362), and \emph{zero-shot} (163), depending on whether they were assigned to more than 50, fewer than 50 but at least one, or no training documents, respectively.

\begin{figure*}[ht]
  \centering
    \includegraphics[width=1\textwidth]{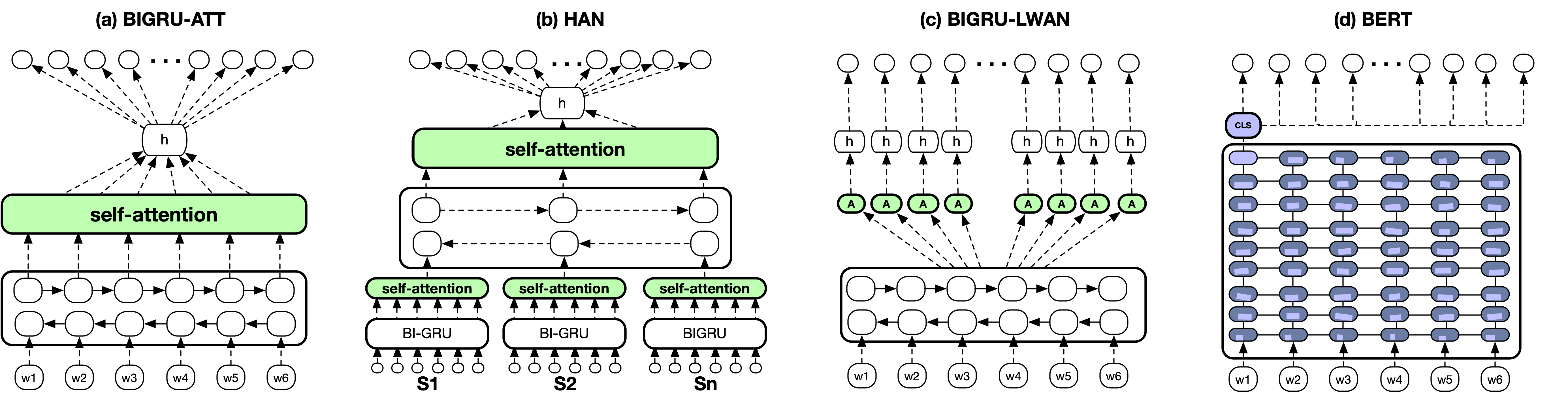}
  \caption{Illustration of (a) \bigruatt, (b) \han, (c) \lwangru, and (d) \bert.}
  \vspace*{-4mm}
  \label{fig:methods}
\end{figure*}

\section{Methods}
\label{sec:methods}

\paragraph{Exact Match, Logistic Regression:} 
A first naive baseline, Exact Match, assigns only labels whose descriptors can be found verbatim in the document. A second one uses Logistic Regression with feature vectors containing \tfidf scores of $n$-grams ($n=1,2,\dots, 5$).

\paragraph{\bigruatt:}

The first neural method is a \textsc{bigru} with self-attention \cite{Xu2015}. Each document is represented as the sequence of its word embeddings, which go through a stack of \textsc{bigru}s (Figure~\ref{fig:methods}a). A document embedding ($h$) is computed as the sum of the resulting context-aware embeddings ($h = \sum_i a_i h_i$), weighted by the self-attention scores ($a_i$), and goes through a dense layer of $L=4,271$ output units with sigmoids, producing $L$ probabilities, one per label.

\paragraph{\han:}

The Hierarchical Attention Network \cite{Yang2016} is a strong baseline for text classification. We use a slightly modified version, where a \bigru with self-attention reads the words of each section, as in \bigruatt but separately per section, producing section embeddings. A second-level \bigru with self-attention reads the section embeddings, producing a single document embedding ($h$) that goes through a similar output layer as in \bigruatt (Figure~\ref{fig:methods}b).

\paragraph{\lwancnn, \lwangru:}

In the original Label-Wise Attention Network (\lwan) of \citet{Mullenbach2018}, called \lwancnn here, the word embeddings of each document are first converted to a sequence of vectors $h_i$ by a \cnn encoder.  A modified version of \lwancnn that we developed, called \lwangru, replaces the \cnn encoder with a \bigru (Figure~\ref{fig:methods}c), which converts the word embeddings into context-sensitive embeddings $h_i$, much as in \bigruatt. Unlike \bigruatt, however, both \lwancnn and \lwangru use $L$ independent attention heads, one per label, generating $L$ document embeddings ($h^{(l)} = \sum_i a_{l,i} h_i$, $l=1, \dots, L$) from the sequence of vectors $h_i$ produced by the \cnn or \bigru encoder, respectively. Each document embedding ($h^{(l)}$) is specialized to predict the corresponding label and goes through a separate dense layer ($L$ dense layers in total) with a sigmoid, to produce the probability of the corresponding label.

\paragraph{\zlwancnn, \zlwangru:}
\citet{Rios2018-2} designed a model similar to \lwancnn, called \zacnn in their work and \zlwancnn here, to deal with rare labels. In \zlwancnn, the attention scores ($a_{l,i}$) and the label probabilities are produced by comparing the $h_i$ vectors that the \cnn encoder produces and the label-specific document embeddings ($h^{(l)}$), respectively, to label embeddings. Each label embedding is the centroid of the pretrained word embeddings of the label's descriptor; consult \citet{Rios2018-2} for further details. By contrast, \lwancnn and \lwangru do not consider the descriptors of the labels. We also experiment with a variant of \zlwancnn that we developed, dubbed \zlwangru, where the \cnn encoder is replaced by a \bigru.

\paragraph{\bert:} 
\bert \citep{bert} is a language model based on Transformers \citep{Vaswani2017} pretrained on large corpora. 
For a new target task, a task-specific layer is added on top of \textsc{bert}. The extra layer is trained jointly with \textsc{bert} by fine-tuning on task-specific data. We add a dense layer on top of \textsc{bert}, with sigmoids, that produces a probability per label. Unfortunately, \bert can currently process texts up to 512 wordpieces, which is too small for the documents of \newdata. Hence, \bert can only be applied to truncated versions of our documents (see below). 

\begin{table*}[ht!]
\centering
{
\footnotesize\addtolength{\tabcolsep}{-2pt}
\begin{tabular}{lccccccccc}
  \hline
  & \multicolumn{3}{c}{\textsc{All Labels}} & \multicolumn{2}{c}{\textsc{Frequent}} & \multicolumn{2}{c}{\textsc{Few}} & \multicolumn{2}{c}{\textsc{Zero}} \\ 
  & $RP@5$ & $nDCG@5$ & Micro-$F1$ & $RP@5$ & $nDCG@5$ & $RP@5$ & $nDCG@5$ & $RP@5$ & $nDCG@5$ \\
  \cline{2-10}
  Exact Match & 0.097 & 0.099 & 0.120 & 0.219 & 0.201 & 0.111 & 0.074 & 0.194 & 0.186 \\
  Logistic Regression & 0.710 & 0.741 & 0.539 & 0.767 & 0.781 & 0.508 & 0.470 & 0.011 & 0.011 \\
  \hline
  \bigruatt & 0.758 & 0.789 & 0.689 & 0.799 & 0.813 & 0.631 & 0.580 & 0.040 & 0.027\\
  \han & 0.746 & 0.778 & 0.680 & 0.789 & 0.805 & 0.597 & 0.544 & 0.051 & 0.034\\
  \hline
 \lwancnn & 0.716 & 0.746 & 0.642 & 0.761 & 0.772 & 0.613 & 0.557 & 0.036  & 0.023 \\
  \lwangru & \textbf{0.766} & \textbf{0.796} & \textbf{0.698} & \textbf{0.805} & \textbf{0.819} & \textbf{0.662} & \textbf{0.618} & 0.029 & 0.019\\
   \hline
  \zlwancnn & 0.684 & 0.717 & 0.618 & 0.730 & 0.745 & 0.495 & 0.454 & 0.321 & 0.264 \\
  \zlwangru & 0.718 & 0.752 & 0.652  & 0.764 & 0.780 & 0.561 & 0.510 & \textbf{0.438} & \textbf{0.345} \\
  \hline
  \hline
 \lwangrulv & 0.775 & 0.804 & 0.711 & 0.815 & 0.828 & 0.656 & 0.612  & 0.034 & 0.024 \\
\hline
\lwangrulv* & 0.770 & 0.796 & 0.709 & 0.811 & 0.825 & 0.641 & 0.600 & 0.047 & 0.030\\
\lwangruelmo* & 0.781 & 0.811 & 0.719 & 0.821 & 0.835 & 0.668 & 0.619 & 0.044 & 0.028\\
\bertbase * & \textbf{0.796} & \textbf{0.823} & \textbf{0.732} & \textbf{0.835} & \textbf{0.846} & \textbf{0.686} & \textbf{0.636} & 0.028 & 0.023\\
\hline
\end{tabular}
}
\caption{Results on \newdata for all, frequent, few-shot, zero-shot labels. Starred methods use the first 512 document tokens; all other methods use full documents. Unless otherwise stated, \glove embeddings are used.}
\vspace*{-4mm}
\label{tab:results}
\end{table*}

\section{Experiments}
\label{sec:experiments}

\paragraph{Evaluation measures:} Common \lmtc evaluation measures are precision ($P@K$) and recall ($R@K$) at the top $K$ predicted labels, averaged over test documents, micro-averaged F1 over all labels, and $nDCG@K$ \cite{Manning2009}. However, $P@K$ and $R@K$ unfairly penalize methods when the gold labels of a document are fewer or more than $K$, respectively. Similar concerns have led to the introduction of $\mathrm{R}\text{-}\mathrm{Precision}$ and $nDCG@K$ in Information Retrieval \cite{Manning2009}, which we believe are also more appropriate for \lmtc. Note, however, that $\mathrm{R}\text{-}\mathrm{Precision}$ requires the number of gold labels per document to be known beforehand, which is unrealistic in practical applications. Therefore we propose using $\mathrm{R}\text{-}\mathrm{Precision}@K$ ($RP@K$), where $K$ is a parameter. This measure is the same as $P@K$ if there are at least $K$ gold labels, otherwise $K$ is reduced to the number of gold labels. 

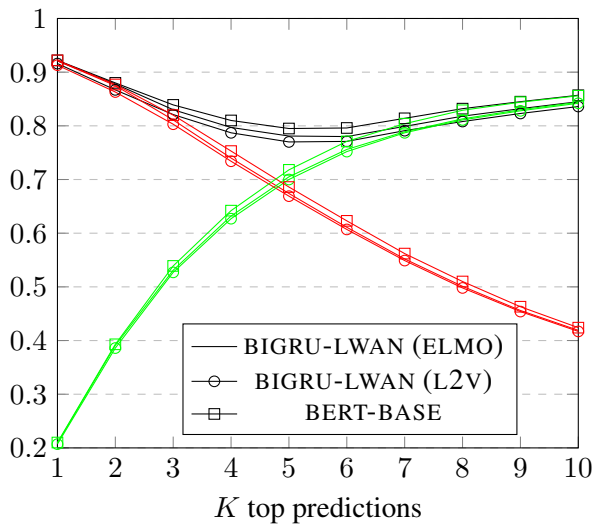
\begin{figure}[ht]
\centering
\begin{tikzpicture}[scale=1.0]
    \begin{axis}[
        xlabel={$K$ top predictions},
        ylabel={},
        xmin=1, xmax=10,
        ymin=0.2, ymax=1,
        xtick={1,2,3,4,5,6,7,8,9,10},
        ytick={0.2, 0.3, 0.4, 0.5, 0.6, 0.7, 0.8, 0.9, 1.0},
        legend style={at={(0.24, 0.03)}, anchor=south west},
        ymajorgrids=true,
        grid style=dashed,
    ]
    
    \addplot[
            color=black,
            ]
            coordinates {
            (1,0.921)(2,0.878)(3,0.830)(4,0.797)(5,0.781)(6,0.780)(7,0.799)(8,0.818)(9,0.832)(10,0.845)};
            \addlegendentry{\lwangru (\elmo)}
            
    \addplot[
            color=black,
            mark=o,
            ]
            coordinates {
            (1,0.916)(2,0.867)(3,0.821)(4,0.787)(5,0.770)(6,0.771)(7,0.791)(8,0.808)(9,0.823)(10,0.836)};
            \addlegendentry{\lwangru (\lawvec)}
            
    \addplot[
            color=black,
            mark=square,
            ]
            coordinates {
            (1,0.922)(2,0.880)(3,0.839)(4,0.810)(5,0.795)(6,0.796)(7,0.814)(8,0.832)(9,0.845)(10,0.857)};
            \addlegendentry{\bertbase}
     
    \addplot[
        color=red,
                ]
        coordinates {
        (1,0.921)(2,0.875)(3,0.811)(4,0.740)(5,0.674)(6,0.611)(7,0.552)(8,0.501)(9,0.456)(10,0.419)};

    \addplot[
        color=green,
                ]
        coordinates {
        (1,0.209)(2,0.391)(3,0.531)(4,0.632)(5,0.705)(6,0.756)(7,0.789)(8,0.813)(9,0.830)(10,0.844)};
    
    \addplot[
        color=red,
        mark=o,
        ]
        coordinates {
        (1,0.913)(2,0.863)(3,0.803)(4,0.734)(5,0.669)(6,0.607)(7,0.549)(8,0.498)(9,0.454)(10,0.417)};

    \addplot[
        color=green,
        mark=o,
        ]
        coordinates {
        (1,0.207)(2,0.386)(3,0.527)(4,0.627)(5,0.700)(6,0.752)(7,0.787)(8,0.811)(9,0.828)(10,0.842)};
    
    \addplot[
        color=red,
        mark=square,
        ]
        coordinates {
        (1,0.922)(2,0.877)(3,0.820)(4,0.753)(5,0.686)(6,0.623)(7,0.562)(8,0.510)(9,0.463)(10,0.424)};

    \addplot[
        color=green,
        mark=square,
        ]
        coordinates {
        (1,0.210)(2,0.393)(3,0.539)(4,0.642)(5,0.718)(6,0.771)(7,0.804)(8,0.829)(9,0.844)(10,0.856)};
     
\end{axis}
\end{tikzpicture}
\caption{$R@K$ (green lines), $P@K$ (red), $RP@K$ (black) of the best methods (\lwangru{s} (\lawvec), \lwangru{s} (\elmo), \bertbase), for $K=1$ to 10. All scores macro-averaged over test documents.}
\label{fig:kappa_comparison}
\end{figure}

Figure~\ref{fig:kappa_comparison} shows $RP@K$ for the three best systems, macro-averaged over test documents. Unlike $P@K$, $RP@K$ does not decline sharply as $K$ increases, because it replaces $K$ by the number of gold labels, when the latter is lower than $K$. 
For $K=1$, $RP@K$ is equivalent to $P@K$, as confirmed by Fig.~\ref{fig:kappa_comparison}. For large values of $K$ that almost always exceed the number of gold labels, $RP@K$ asymptotically approaches $R@K$, as also confirmed by Fig.~\ref{fig:kappa_comparison}.\footnote{See Appendix~\ref{app:evaluation} for a more detailed discussion on the evaluation measures.} In our dataset, there are 5.07 labels per document, hence  $K=5$ is reasonable.\footnote{Evaluating at other values of $K$ lead to similar conclusions (see Fig.~\ref{fig:kappa_comparison} and Appendix~\ref{app:experiments}).}

\paragraph{Setup:} 
Hyper-parameters are tuned using the \textsc{hyperopt} library selecting the values with the best loss on development data.\footnote{We implemented all methods in \textsc{keras} (\url{https://keras.io/}). Code available at \url{https://github.com/iliaschalkidis/lmtc-eurlex57k.git}. See Appendix~\ref{app:hyperopt} for details on hyper-parameter tuning.} For the best hyper-parameter values, we perform five runs and report mean scores on test data. For statistical significance tests, we take the run of each method with the best performance on development data, and perform two-tailed approximate randomization tests \cite{Dror2018} on test data.\footnote{We perform 10k iterations, randomly swapping in each iteration the responses (sets of returned labels) of the two compared systems for 50\% of the test documents.} Unless otherwise stated, we used 200-D pretrained \glove embeddings \cite{pennington2014glove}.

\paragraph{Full documents:}
\label{sec:overall}
The first five horizontal zones of Table~\ref{tab:results} report results for full documents. The naive baselines are weak, as expected. Interestingly, for all, frequent, and even few-shot labels, the generic \bigruatt performs better than \lwancnn, which was designed for \lmtc. \han also performs better than \lwancnn for all and frequent labels. However, replacing the \cnn encoder of \lwancnn with a \bigru (\lwangru) leads to the best results, indicating that the main weakness of \lwancnn is its vanilla \cnn encoder.

The zero-shot versions of \lwancnn and \lwangru outperform all other  methods on zero-shot labels (Table~\ref{tab:results}), in line with the findings of \citet{Rios2018-2}, because they exploit label descriptors,
but more importantly because they have a component that uses prior knowledge as is (i.e., label embeddings are frozen). Exact Match also performs better on zero-shot labels, for the same reason (i.e., the prior knowledge is intact). \lwangru, however, is still the best method in few-shot learning. All the differences between the best (bold) and other methods in Table~\ref{tab:results} are statistically significant ($p < 0.01$).

Table~\ref{tab:embs} shows that using \wordvec embeddings trained on legal texts (\lawvec) \cite{Chalkidis2018} or \elmo embeddings \cite{Peters2018} trained on generic texts further improve the performance of \lwangru. 

\paragraph{Document zones:} 
Table~\ref{tab:zones} compares the performance of \lwangru on the development set for different combinations of document zones (Section~\ref{sec:dataset}): \emph{header} (\emph{H}), \emph{recitals}  (\emph{R}), \emph{main body} (\emph{MB}), full text. Surprisingly \emph{H+R} leads to almost the same results as full documents,\footnote{The approximate randomization tests detected no statistically significant difference in this case ($p = 0.20$).} indicating that \emph{H+R} provides most of the information needed to assign \eurovoc labels.

\begin{table}[ht!]
\centering
\footnotesize\addtolength{\tabcolsep}{-2pt}
\begin{tabular}{lccc}
  & $RP@5$ & $nDCG@5$ & Micro-$F1$ \\
\cline{2-4}
\glove & 0.766 & 0.796 & 0.698 \\
\lawvec & 0.775 & 0.804 & 0.711 \\
\glove + \elmo & 0.777 & 0.808 & 0.714\\
\lawvec + \elmo & \textbf{0.781} & \textbf{0.811} & \textbf{0.719}\\
\end{tabular}
\caption{\lwangru with \glove, \lawvec, \elmo.}
\label{tab:embs}
\end{table}
\vspace{-0.5em}
\begin{table}[ht!]
\centering
\footnotesize\addtolength{\tabcolsep}{-2pt}
\begin{tabular}{lcccc}
  & $\mu_{words}$ & $RP@5$ & $nDCG@5$ & Micro-$F1$ \\
\cline{2-5}
\emph{H} & 43 & 0.747 & 0.782 & 0.688 \\
\emph{R}  & 317 & 0.734 & 0.765 & 0.669 \\
\emph{H+R} & 360 & \underline{0.765} & \underline{0.796} & \underline{0.701} \\
\emph{MB} & 187 & 0.643 & 0.674 & 0.590 \\
\emph{Full} & 727 & \textbf{0.766} & \textbf{0.797} & \textbf{0.702} \\
\end{tabular}
\caption{\lwangru with different document zones.}
\vspace*{-4mm}
\label{tab:zones}
\end{table}

\paragraph{First 512 tokens:} Given that \emph{H+R} contains enough information and is shorter than 500 tokens in 83\% of our dataset's documents, we also apply \bert to the first 512 tokens of each document (truncated to \bert's max.\ length), comparing to \lwangru also operating on the first 512 tokens. Table~\ref{tab:results} (bottom zone) shows that \bert outperforms all other methods, even though it considers only the first 512 tokens. It fails, however, in zero-shot learning, since it does not have a component that exploits prior knowledge as is (i.e., all the components are fine-tuned on training data).

\section{Limitations and Future Work}

One major limitation of the investigated methods is that they are unsuitable for \emph{Extreme} Multi-Label Text Classification where there are hundreds of thousands of labels \cite{Liu2017,Zhang2018,Wydmuch2018}, as opposed to the \lmtc setting of our work where the labels are in the order of thousands. We leave the investigation of methods for extremely large label sets for future work. Moreover, \textsc{rnn} (and \gru) based methods have high computational cost, especially for long documents. We plan to investigate more computationally efficient methods, e.g., dilated \cnn{s} \cite{KalchbrennerESO16} and Transformers \cite{Vaswani2017, Dai2019}. We also plan to experiment with hierarchical flavors of \bert to surpass its length limitations. Furthermore, experimenting with more datasets e.g., \rcv, Amazon-13K, Wiki-30K, \mimiciii will allow us to confirm our conclusions in different domains. Finally, we plan to investigate Generalized Zero-Shot Learning \cite{Liu2018}. 

\section*{Acknowledgements}
This work was partly supported by the Research Center of the Athens University of Economics and Business.

\vspace{-3mm}

\bibliography{acl2019}
\bibliographystyle{acl_natbib}

\appendix
\section*{Appendix}

\section{\newdata statistics}
\label{app:dataset}

\begin{table*}[h]
\centering
{
\small
\begin{tabular}{lccccccc}
  \hline
  Hyper parameters & \bigruatt & \han & \lwancnn & \lwangru & \zacnn* & \zagru* & \bertbase+ \\
  $N_l\in[1, 2]$ & 1 & (1,1) & 1 & 1 & 1 & 1 & 12 \\
  $HU\in[200, 300, 400]$ & 300 & (300,300) & 200 & 300 & 200 & 100 & 768 \\
  $D_d\in[0.1, 0.2, \dots, 0.5]$ & 0.2 & 0.3 & 0.1 & 0.4 & 0.1 & 0.1 & 0.1 \\
  $D_{we}\in[0.00, 0.01, 0.02]$ & 0.02 & 0.02 & 0.01 & 0.00 & 0.00 & 0.00 & 0.00 \\
  $BS\in[8, 12, 16]$ & 12 & 16 & 12 & 16 & 16 & 16 & 8 \\
  \hline
 \end{tabular}
 }
  \caption{Best hyper parameters for neural methods. $N_l$: number of layers, $HU$: hidden units size, $D_d$: dropout rate across dimensions, $D_{we}$: dropout rate of word embeddings, $BS$: batch size. * Hidden units size is fixed to word embedding dimensionality, + $N_l$, $HU$ are fixed from the pre-trained model. Dropout rate fixed as suggested by Devlin et al. (2018).}
\label{tab:parameters}
\end{table*}

\begin{figure}[t!]
  \centering
    \includegraphics[width=\columnwidth]{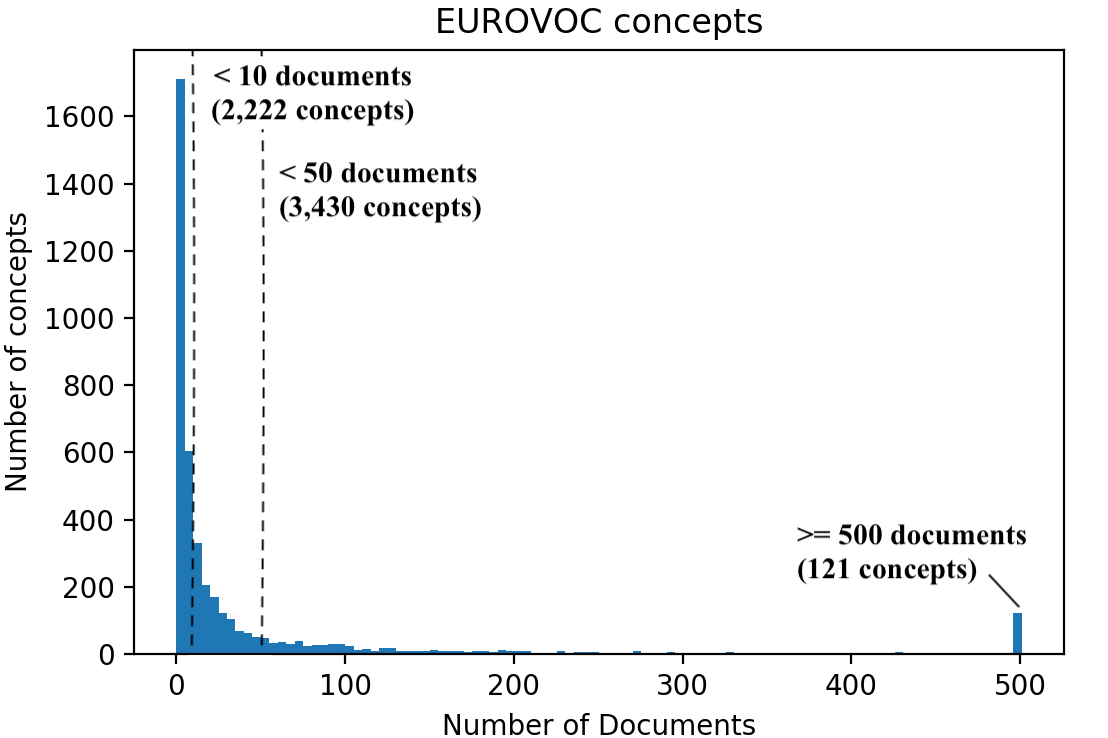}
  \caption{Distribution of \eurovoc concepts across \newdata documents}
  \label{fig:histogram}
\end{figure}

Figure~\ref{fig:histogram} shows the distribution of labels across \newdata documents. From the 7k labels fewer than 50\% appear in more than 10 documents. Such an aggressive Zipfian distribution has also been noted in medical code predictions \cite{Rios2018-2}, where such thesauri are used to classify documents, demonstrating the practical importance of few-shot and zero-shot learning. 

\section{Hyper-paramater tuning}
\label{app:hyperopt}
Table~\ref{tab:parameters} shows the best hyper-parameters returned by \textsc{hyperopt}. Concerning \bert, we set the dropout rate and learning rate to 0.1 and 5e-5, respectively, as suggested by \citet{bert}, while batch size was set to 8 due to \gpu memory limitations. Finally, we noticed that the model did not converge in the fourth epoch, as suggested by \citet{bert}. Thus we used early-stopping with no patience and trained the model for eight to nine epochs on average among the five runs. 

\section{Evaluation Measures}

\begin{table*}[ht!]
\centering
{
\footnotesize\addtolength{\tabcolsep}{-2pt}
\begin{tabular}{lcccccccccccc}
  \hline
  & \multicolumn{3}{c}{\textsc{Overall}} & \multicolumn{3}{c}{\textsc{Frequent}} & \multicolumn{3}{c}{\textsc{Few}} & \multicolumn{3}{c}{\textsc{Zero}} \\ 
  & @1 & @5 & @10 & @1 & @5 & @10 & @1 & @5 & @10 & @1 & @5 & @10 \\
  \cline{2-13}
  Exact Match & 0.131 & 0.084 & 0.080 & 0.194 & 0.166 & 0.141 & 0.037 & 0.037 & 0.036 & 0.178 & 0.042 & 0.022 \\
  Logistic Regression & 0.861 & 0.613 & 0.378 & 0.864 & 0.604 & 0.368 & 0.458 & 0.169 & 0.094 & 0.011 & 0.002 & 0.002 \\
  \hline
  \bigruatt & 0.899 & 0.654 & 0.407 & 0.893 & 0.627 & 0.382 & 0.551 & 0.212 & 0.121 & 0.015 & 0.008 & 0.007 \\
  \han & 0.894 & 0.643 & 0.401 & 0.889 & 0.620 & 0.378 & 0.510 & 0.199 & 0.114 & 0.020 & 0.011 & 0.008 \\
  \hline
 \lwancnn & 0.853 & 0.617 & 0.395 & 0.849 & 0.596 & 0.374 & 0.521 & 0.204 & 0.117 & 0.011 & 0.007 & 0.007 \\
  \lwangru & \underline{0.907} & \underline{0.661} & \underline{0.414} & \underline{0.900} & \underline{0.631} & \underline{0.387} & \underline{0.599} & \underline{0.222} & \underline{0.124} & 0.011 & 0.006 & 0.006 \\
   \hline
  \zlwancnn & 0.842 & 0.589 & 0.371 & 0.837 & 0.572 & 0.355 & 0.447 & 0.164 & 0.094 & \underline{0.202} & \underline{0.069} & \underline{0.040} \\
  \zlwangru & 0.874 & 0.619 & 0.386 & 0.867 & 0.599 & 0.367 & 0.488 & 0.184 & 0.107 & \textbf{0.247} & \textbf{0.093} & \textbf{0.057} \\
  \hline\hline
   \lwangrulv & 0.913 & 0.669 & 0.417 & 0.905 & 0.639 & 0.390 & 0.593 & 0.219 & 0.122 & 0.013 & 0.007 & 0.008 \\
\hline
\lwangrulv* & 0.915 & 0.664 & 0.413 & 0.905 & 0.637 & 0.387 & 0.586 & 0.214 & 0.120 & 0.013 & 0.010 & 0.010 \\
\lwangruelmo* & 0.921 & 0.674 & 0.419 & 0.912 & 0.644 & 0.391 & 0.595 & 0.226 & 0.127 & 0.011 & 0.009 & 0.007 \\
  \bertbase* & \textbf{0.922} & \textbf{0.687} & \textbf{0.424} & \textbf{0.914} & \textbf{0.656} & \textbf{0.394} & \textbf{0.611} & \textbf{0.229} & \textbf{0.129} & 0.019 & 0.006 & 0.007 \\
  \hline
\end{tabular}
}
\caption{$P@1$, $P@5$ and $P@10$ results on \newdata for all, frequent, few-shot, zero-shot labels. Starred methods use the first 512 document tokens; all other methods use full documents. Unless otherwise stated, \glove embeddings are used.}
\label{tab:presults}
\end{table*}

\label{app:evaluation}
The macro-averaged versions of $R@K$ and $P@K$ are defined as follows:
\begin{flalign}
R@K &= \frac{1}{T} \sum^T_{t=1}\frac{S_t(K)}{R_t}\label{eq:r@k}\\
P@K &= \frac{1}{T} \sum^T_{t=1}\frac{S_t(K)}{K}\label{eq:p@k}
\end{flalign}

\noindent where $T$ is the total number of test documents, $K$ is the number of labels to be selected per document, $S_t(K)$ is the number of correct labels among those ranked as top $K$ for the $t$-th document, and $R_t$ is the number of gold labels for each document.
Although these measures are widely used in \lmtc, we question their appropriateness for the following reasons:

\begin{enumerate}[leftmargin=1em]
    \item $R@K$ leads to excessive penalization when documents have more than $K$ gold labels. For example, evaluating at $K=1$ for a single document with 5 gold labels returns $R@1=0.20$, if the system managed to return a correct label.
    The system is penalized, even though it was not allowed to return more than one label.
    \item $P@K$ does the same for documents with fewer than $K$ gold labels. For example, evaluating at $K=5$ for a single document with a single gold label returns $P@1=0.20$.
    \item Both measures over- or under-estimate performance on documents whose number of gold labels largely diverges from $K$. 
    This is clearly illustrated in Figure~\ref{fig:kappa_comparison} of the main article.
    \item Because of these drawbacks, both measures do not correctly single out the best methods.
\end{enumerate}

\begin{figure}[h!]
\centering
\begin{tikzpicture}[scale=0.9]
    \begin{axis}[
        xlabel={Number of labels per document},
        xmin=0, xmax=27,
        ymin=0, ymax=1.1,
        xtick={1,5,10,15,20,26},
        ytick={0.0, 0.1, 0.2, 0.3, 0.4, 0.5, 0.6, 0.7, 0.8, 0.9, 1.0},
        legend style={at={(0.98, 0.55)},anchor=east},
        ymajorgrids=true,
        grid style=dashed,
    ]
    
    \addplot[
        color=blue,
        mark=otimes,
        ]
        coordinates {
        (1, 0.00539)(2, 0.04834)(3, 0.13683)(4, 0.17454)(5, 0.21180)(6, 0.28330)(7, 0.07302)(8, 0.03807)(9, 0.01528)(10, 0.00786)(11, 0.00291)(12, 0.00130)(13, 0.00060)(14, 0.00028)(15, 0.00021)(16, 0.00012)(17, 0.00009)(18, 0.00004)(19, 0.00002)(20, 0.00000)(21, 0.00000)(22, 0.00000)(23, 0.00000)(24, 0.00000)(25, 0.00000)(26, 0.00002)};
        \addlegendentry{Probability distribution}
    
    \addplot[
        color=orange,
        mark=square,
        ]
        coordinates {
        (1, 0.00539)(2, 0.05372)(3, 0.19055)(4, 0.36509)(5, 0.57689)(6, 0.86019)(7, 0.93321)(8, 0.97128)(9, 0.98656)(10, 0.99442)(11, 0.99733)(12, 0.99863)(13, 0.99923)(14, 0.99951)(15, 0.99972)(16, 0.99984)(17, 0.99993)(18, 0.99996)(19, 0.99998)(20, 0.99998)(21, 0.99998)(22, 0.99998)(23, 0.99998)(24, 0.99998)(25, 0.99998)(26, 1.00000)};
        \addlegendentry{Cumulative distribution}
     
    \end{axis}
    \end{tikzpicture}
\caption{Distribution of number of labels per document in \newdata.}
\label{fig:labels-density}
\end{figure}
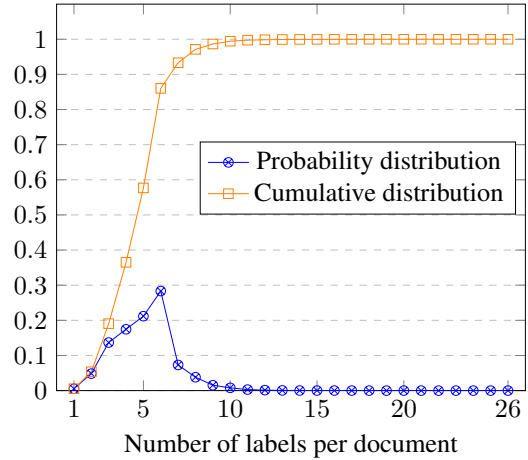

Based on the above arguments, we believe that R-Precision@K ($RP@K$) and $nDCG@K$ lead to a more informative and fair evaluation. Both measures adjust to the number of gold labels per document, without over- or under-estimating performance when documents have few or many gold labels.
The macro-averaged versions of the two measures are defined as follows:
\vspace{-3mm}
\begin{flalign}
RP@K &= \frac{1}{T} \sum^T_{t=1}\frac{S_t(K)}{\min{(K,R_t)}}\\
nDCG@K &= \frac{1}{T} \sum^T_{t=1} \sum^K_{k=1} \frac{2^{S_t(k)}-1}{\log{(1+k)}} 
\end{flalign}
Again, $T$ is the total number of test documents, $K$ is the number of labels to be selected, $S_t(K)$ is the number of correct labels among those ranked as top $K$ for the $t$-th document, and $R_t$ is the number of gold labels for each document. In the main article we report results for $K=5$. The reason is that the majority of the documents of \newdata (57.7\%) have at most 5 labels.
The detailed distributions can be seen in Figure~\ref{fig:labels-density}.

\section{Experimental Results}
\label{app:experiments}
In Tables~\ref{tab:presults}--\ref{tab:ngcgresults}, we present additional results
for the main measures used across the \lmtc literature ($P@K$, $R@K$, $RP@K$, $nDGC@K$).
\vspace{-2mm}

\begin{table*}[h!]
\centering
{
\footnotesize\addtolength{\tabcolsep}{-2pt}
\begin{tabular}{lcccccccccccc}
  \hline
  & \multicolumn{3}{c}{\textsc{Overall}} & \multicolumn{3}{c}{\textsc{Frequent}} & \multicolumn{3}{c}{\textsc{Few}} & \multicolumn{3}{c}{\textsc{Zero}} \\ 
  & @1 & @5 & @10 & @1 & @5 & @10 & @1 & @5 & @10 & @1 & @5 & @10 \\
  \cline{2-13}
  Exact Match & 0.026 & 0.087 & 0.168 & 0.045 & 0.207 & 0.344 & 0.022 & 0.111 & 0.214 & 0.161 & 0.194 & 0.206 \\
  Logistic Regression & 0.195 & 0.641 & 0.764 & 0.234 & 0.719 & 0.845 & 0.313 & 0.507 & 0.560 & 0.011 & 0.011 & 0.022 \\
  \hline
  \bigruatt & 0.204 & 0.685 & 0.824 & 0.242 & 0.749 & 0.880 & 0.382 & 0.629 & 0.703 & 0.015 & 0.040 & 0.062 \\
  \han & 0.203 & 0.675 & 0.811 & 0.241 & 0.740 & 0.871 & 0.355 & 0.596 & 0.673 & 0.018 & 0.051 & 0.079 \\
  \hline
 \lwancnn & 0.193 & 0.647 & 0.800 & 0.229 & 0.713 & 0.862 & 0.360 & 0.612 & 0.681 & 0.011 & 0.036 & 0.061 \\
  \lwangru & 0.205 & 0.692 & 0.836 & 0.243 & 0.755 & 0.891 & 0.420 & 0.661 & 0.725 & 0.011 & 0.029 & 0.060 \\
   \hline
  \zlwancnn & 0.189 & 0.617 & 0.752 & 0.223 & 0.683 & 0.820 & 0.300 & 0.494 & 0.556 & 0.189 & 0.321 & 0.376 \\
  \zlwangru & 0.197 & 0.648 & 0.782 & 0.232 & 0.716 & 0.847 & 0.335 & 0.560 & 0.635 & \textbf{0.231} & \textbf{0.438} & \textbf{0.531} \\
  \hline\hline
   \lwangrulv & 0.207 & 0.700 & 0.842 & 0.246 & 0.764 & 0.898 & 0.414 & 0.655 & 0.716 & 0.012 & 0.034 & 0.066 \\
\hline
\lwangrulv* & 0.207 & 0.696 & 0.835 & 0.245 & 0.760 & 0.891 & 0.409 & 0.640 & 0.707 & 0.013 & 0.047 & 0.084 \\
\lwangruelmo* & 0.208 & 0.705 & 0.844 & 0.249 & 0.770 & 0.900 & 0.410 & 0.667 & 0.732 & 0.011 & 0.044 & 0.061 \\
  \bertbase* & \textbf{0.209} & \textbf{0.719} & \textbf{0.855} & \textbf{0.250} & \textbf{0.784} & \textbf{0.908} & \textbf{0.428} & \textbf{0.684} & \textbf{0.752} & 0.018 & 0.028 & 0.068 \\
  \hline
\end{tabular}
}
\caption{$R@1$, $R@5$ and $R@10$ results on \newdata for all, frequent, few-shot, zero-shot labels. Starred methods use the first 512 document tokens; all other methods use full documents. Unless otherwise stated, \glove embeddings are used.}
\label{tab:rresults}
\end{table*}

\vspace{-2mm}

\begin{table*}[h!]
\centering
{
\footnotesize\addtolength{\tabcolsep}{-2pt}
\begin{tabular}{lcccccccccccc}
  \hline
  & \multicolumn{3}{c}{\textsc{Overall}} & \multicolumn{3}{c}{\textsc{Frequent}} & \multicolumn{3}{c}{\textsc{Few}} & \multicolumn{3}{c}{\textsc{Zero}} \\ 
  & @1 & @5 & @10 & @1 & @5 & @10 & @1 & @5 & @10 & @1 & @5 & @10 \\
  \cline{2-13}
  Exact Match & 0.131 & 0.097 & 0.168 & 0.194 & 0.219 & 0.344 & 0.037 & 0.111 & 0.214 & 0.178 & 0.194 & 0.206 \\
  Logistic Regression & 0.861 & 0.710 & 0.765 & 0.864 & 0.767 & 0.846 & 0.458 & 0.508 & 0.560 & 0.011 & 0.011 & 0.022 \\
  \hline
  \bigruatt & 0.899 & 0.758 & 0.824 & 0.893 & 0.799 & 0.880 & 0.551 & 0.631 & 0.703 & 0.015 & 0.040 & 0.062 \\
  \han{s} & 0.894 & 0.746 & 0.811 & 0.889 & 0.789 & 0.872 & 0.510 & 0.597 & 0.673 & 0.020 & 0.051 & 0.079 \\
  \hline
 \lwancnn & 0.853 & 0.716 & 0.801 & 0.849 & 0.761 & 0.862 & 0.521 & 0.613 & 0.681 & 0.011 & 0.036 & 0.061 \\
  \lwangru & \underline{0.907} & \underline{0.766} & \underline{0.836} & \underline{0.900} & \underline{0.805} & \underline{0.891} & \underline{0.599} & \underline{0.662} & \underline{0.725} & 0.011 & 0.029 & 0.060 \\
   \hline
  \zlwancnn & 0.842 & 0.684 & 0.753 & 0.837 & 0.730 & 0.820 & 0.447 & 0.495 & 0.556 & \underline{0.202} & \underline{0.321} & \underline{0.376} \\
  \zlwangru & 0.874 & 0.718 & 0.782 & 0.867 & 0.764 & 0.847 & 0.488 & 0.561 & 0.635 & \textbf{0.247} & \textbf{0.438} & \textbf{0.531} \\
  \hline\hline
   \lwangrulv & 0.913 & 0.775 & 0.842 & 0.905 & 0.815 & 0.898 & 0.593 & 0.657 & 0.716 & 0.013 & 0.034 & 0.066 \\
\hline
\lwangrulv* & 0.915 & 0.770 & 0.836 & 0.905 & 0.811 & 0.891 & 0.586 & 0.641 & 0.707 & 0.013 & 0.047 & 0.084 \\
\lwangruelmo* & 0.921 & 0.781 & 0.845 & 0.912 & 0.821 & 0.901 & 0.595 & 0.668 & 0.732 & 0.011 & 0.044 & 0.061 \\
  \bertbase* & \textbf{0.922} & \textbf{0.796} & \textbf{0.856} & \textbf{0.914} & \textbf{0.835} & \textbf{0.908} & \textbf{0.611} & \textbf{0.686} & \textbf{0.752} & 0.019 & 0.028 & 0.068 \\
  \hline
\end{tabular}
}
\caption{$RP@1$, $RP@5$ and $RP@10$ results on \newdata for all, frequent, few-shot, zero-shot labels. Starred methods use the first 512 document tokens; all other methods use full documents. Unless otherwise stated, \glove embeddings are used.}
\label{tab:rpresults}
\end{table*}

\vspace{-2mm}

\begin{table*}[h!]
\centering
{
\footnotesize\addtolength{\tabcolsep}{-2pt}
\begin{tabular}{lcccccccccccc}
  \hline
  & \multicolumn{3}{c}{\textsc{Overall}} & \multicolumn{3}{c}{\textsc{Frequent}} & \multicolumn{3}{c}{\textsc{Few}} & \multicolumn{3}{c}{\textsc{Zero}} \\ 
  & @1 & @5 & @10 & @1 & @5 & @10 & @1 & @5 & @10 & @1 & @5 & @10 \\
  \cline{2-13}
  Exact Match & 0.131 & 0.099 & 0.134 & 0.194 & 0.201 & 0.262 & 0.037 & 0.074 & 0.112 & 0.178 & 0.186 & 0.189 \\
  Logistic Regression & 0.861 & 0.741 & 0.766 & 0.864 & 0.781 & 0.819 & 0.458 & 0.470 & 0.489 & 0.011 & 0.011 & 0.014 \\
  \hline
  \bigruatt & 0.899 & 0.789 & 0.819 & 0.893 & 0.813 & 0.853 & 0.551 & 0.580 & 0.608 & 0.015 & 0.027 & 0.034 \\
  \han & 0.894 & 0.778 & 0.808 & 0.889 & 0.805 & 0.845 & 0.510 & 0.544 & 0.573 & 0.020 & 0.034 & 0.043 \\
  \hline
 \lwancnn & 0.853 & 0.746 & 0.786 & 0.849 & 0.772 & 0.822 & 0.521 & 0.557 & 0.583 & 0.011 & 0.023 & 0.032 \\
  \lwangru & 0.907 & 0.796 & 0.829 & 0.900 & 0.819 & 0.861 & 0.599 & 0.618 & 0.643 & 0.011 & 0.019 & 0.029 \\
   \hline
  \zlwancnn & 0.842 & 0.717 & 0.749 & 0.837 & 0.745 & 0.789 & 0.447 & 0.454 & 0.478 & 0.202 & 0.264 & 0.281 \\
  \zlwangru & 0.874 & 0.752 & 0.781 & 0.867 & 0.780 & 0.819 & 0.488 & 0.510 & 0.539 & \textbf{0.247} & \textbf{0.345} & \textbf{0.375} \\
  \hline\hline
   \lwangrulv & 0.913 & 0.804 & 0.836 & 0.905 & 0.828 & 0.869 & 0.593 & 0.612 & 0.635 & 0.013 & 0.024 & 0.035 \\
\hline
\lwangrulv* & 0.915 & 0.801 & 0.832 & 0.905 & 0.825 & 0.864 & 0.586 & 0.600 & 0.625 & 0.013 & 0.030 & 0.042 \\
\lwangruelmo* & 0.921 & 0.811 & 0.841 & 0.912 & 0.835 & 0.874 & 0.595 & 0.619 & 0.643 & 0.011 & 0.028 & 0.034 \\
  \bertbase* & \textbf{0.922} & \textbf{0.823} & \textbf{0.851} & \textbf{0.914} & \textbf{0.846} & \textbf{0.882} & \textbf{0.611} & \textbf{0.636} & \textbf{0.662} & 0.019 & 0.023 & 0.036 \\
  \hline
\end{tabular}
}
\caption{$nDCG@1$, $nDCG@5$ and $nDCG@10$ results on \newdata for all, frequent, few-shot, zero-shot labels. Starred methods use the first 512 document tokens; all other methods use full documents. Unless otherwise stated, \glove embeddings are used.}
\label{tab:ngcgresults}
\end{table*}
\end{document}